\documentclass[runningheads]{llncs}
\usepackage[T1]{fontenc}
\usepackage{graphicx}
\usepackage{cite} 

\usepackage{pifont}
\newcommand{\cmark}{\ding{51}} 
\usepackage{amsmath}

\usepackage{subcaption} 
\usepackage{xcolor} 
\begin{document}
\title{CollideNet: Hierarchical Multi-scale Video Representation Learning with Disentanglement for Time-To-Collision Forecasting}
\titlerunning{CollideNet: Hierarchical Multi-scale Video Learning for TTC Forecasting}
%
\author{Nishq Poorav Desai\inst{1}\orcidID{0000-0002-4185-266X} \and
Michael Greenspan\inst{1}\orcidID{0000-0001-6054-8770} \and
Ali Etemad\inst{1}\orcidID{0000-0001-7128-0220}}
\authorrunning{N.P. Desai et al.}
%
\institute{Queen's University, Canada \\
\email{\{n.desai, ali.etemad, michael.greenspan\}@queensu.ca}}
\maketitle              
\begin{abstract}
Time-to-Collision (TTC) forecasting is a critical task in collision prevention, requiring precise temporal prediction and comprehending both local and global patterns encapsulated in a video, both spatially and temporally. To address the multi-scale nature of video, we introduce a novel spatiotemporal hierarchical transformer-based architecture called CollideNet,  specifically catered for effective TTC forecasting. In the spatial stream, CollideNet aggregates information for each video frame simultaneously at multiple resolutions. In the temporal stream, along with multi-scale feature encoding, CollideNet also disentangles the non-stationarity, trend, and seasonality components. Our method achieves state-of-the-art performance in comparison to prior works on three commonly used public datasets, setting a new state-of-the-art by a considerable margin.
We  
conduct cross-dataset evaluations to analyze the generalization capabilities of our method, and visualize the effects of disentanglement of the trend and seasonality components of the video data. 
We release our code at \url{https://github.com/DeSinister/CollideNet/}.

\keywords{Time-to-Collision Forecasting \and Hierarchical Spatiotemporal Modeling \and Disentangled Representation Learning.}
\end{abstract}

\section{Introduction}
\label{sec:intro}

In recent years, automated collision warning systems in the automotive industry have achieved significant advancements toward safer autonomous driving, preventing accidents and enhancing real-time decision-making in advanced driver assistance systems (ADAS)~\cite{nhtsa2024}. To brake effectively, it is necessary to accurately estimate the time for a potential collision, known as the Time-To-Collision (TTC), which has also been used to alert drivers and control autonomous vehicles to decelerate~\cite{janssen1994ttc}. A study conducted by the Insurance Institute for Highway Safety (IIHS) reported that forward collision warnings reduced the rear-end crash rate per vehicle mile traveled by 22\%~\cite{teoh2021effectiveness}. Research by Daimler-Benz~\cite{ntsb2001} found that a 1.5-second warning could prevent up to 90\% of collisions, while even a brief 0.5-second warning could  avoid 60\% of potential collisions.

\begin{figure}[t]
      \centering
      \includegraphics[width=\linewidth]{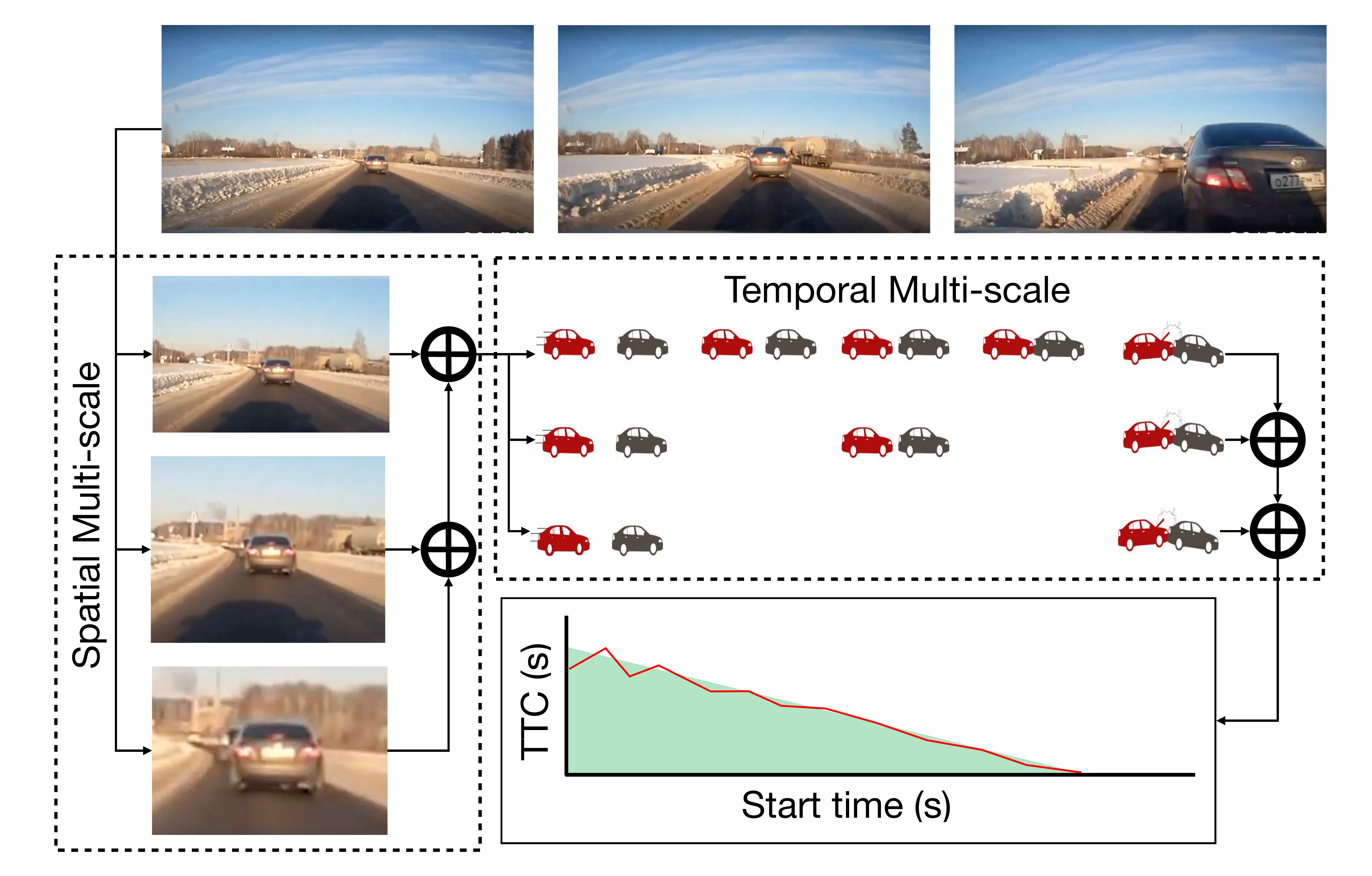}
      \caption{TTC forecasting task using multi-scale spatial and temporal information.}
      \label{fig:banner}
    
\end{figure}

Accurate TTC forecasting is challenging because models must interpret both short-range details (e.g., nearby objects and sudden events) and long-range context (e.g., overall traffic flow). They must also handle limited visibility and occlusions, which makes it necessary to predict how objects and actions change in uncertain environments, making this task far more complex than basic recognition or tracking. One approach to consider the long- and short-range spatial and temporal dependencies is to capture features at multiple resolutions and combine them effectively via a multi-scale architecture~\cite{Wang2023MICNML}. 
Existing TTC forecasting methods often rely on convolutional spatial filters, which provide spatial multi-scale receptive fields. However, these filters typically have small receptive fields and struggle to integrate global and local information effectively~\cite{dosovitskiy2021an,Qiu2019LearningSR}. Moreover, they do not incorporate a multi-scale approach for the temporal dimension. On the other hand, 
attention-based models such as transformers, which are commonly used to capture \textit{temporal} patterns~\cite{anjum2023spatio,anjum2023learning,desai2025cyclecrash}, 
are not inherently multi-scale and face challenges in learning long-range dependencies in video data. This is due to the quadratic space and time complexity of self-attention, which hinders its scalability~\cite{Nagar2024SEMASA, Yoo2023TowardsEG, Qin2023FactorizationVT}.
Therefore, there is a clear need for multi-scale models that integrate both short-term and long-term feature extraction in spatial and temporal dimensions to enhance TTC forecasting.

To address these limitations,  we propose a novel hierarchical multi-scale Transformer-based method called \textit{CollideNet}, with the aim of capturing both 
local and global spatial and temporal features in traffic videos for TTC forecasting (see Figure~\ref{fig:banner}). To overcome the issue of occluded collisions or collisions occurring in localized regions where background noise may interfere with model learning, we propose an efficient module that leverages low-rank decomposition techniques. Inspired by prior works~\cite{6843945,7406471,8265430,Grosek2014DynamicMD} that separate foreground and background for traffic-related tasks, we enhance this approach by modeling long-range background and short-range foreground dynamics using series decomposition~\cite{rb1990stl,Anderson1976TimeSeries2E}.
The series decomposition is performed using a moving average with a predefined window size $k$ to extract the trend component, while the remaining time-series forms the seasonal component.
Unlike existing methods that operate directly on spatial dimensions, which leads to high computational costs, our method applies these decompositions on spatially encoded frames, drastically reducing computational overhead while disentangling seasonality (short-range) and trend (long-range) patterns.

In the temporal dimension, the standard point-wise self-attention struggles with capturing long-range predictive dependencies and suffers from a quadratic complexity. We integrate a segment-wise correlation-based attention instead that divides the time-series into segments and applies segment-wise correlation, inspired by \cite{du2023preformer}. This approach not only reduces computational complexity but also significantly improves the model's ability to capture predictive dependencies across multiple scales. 
Furthermore, to aid with long-range temporal modeling in transformers, we leverage the disentanglement of spatially encoded features along stationarity, building on ideas from \cite{du2023preformer,liu2022non,taylor2018forecasting,oreshkin2019n} for more robust and efficient representations. To demonstrate the effectiveness of our proposed method, we conduct various experiments on 3 public datasets, the Dashcam Accident Dataset (DAD) \cite{chan2016anticipating}, the Car Crash Dataset (CCD) \cite{BaoMM2020}, and the Detection of Traffic Anomaly Dataset (DoTA) \cite{yao2022dota}. Additionally, we perform thorough ablation studies to determine the impact of each key component of our method, as well as sensitivity analyses on different hyperparameters.

The main contributions of the paper are:
    (\textbf{1}) We present a novel two-stream
    hierarchical multi-scale transformer architecture in CollideNet, specifically designed to cater to the hierarchical spatial and temporal features for TTC forecasting in videos. 
    (\textbf{2}) For the first time in the context of TTC, we model the disentanglement of temporal patterns, namely non-stationarity, trend, and seasonality for temporal encoding, to enhance forecasting. We further notice that the enhancement in forecasting from disentanglement is compounded when used with the multi-scale architecture.
    (\textbf{3}) Our method achieves state-of-the-art performance on three datasets, especially outperforming previous methods by a large margin in the CCD dataset. Furthermore, we also demonstrate the generalizability of our model via cross-dataset evaluation. To enable rapid reproducibility of these results and to contribute to the area, we release our code at \url{https://github.com/DeSinister/CollideNet/}.

\section{Related work}
In this section, we review prior work on TTC forecasting. We also discuss collision anticipation and detection, which are closely related but are comparatively easier tasks, as they often rely on detection methods that do not require prediction of future events and are typically framed as classification problems. In contrast, TTC forecasting is a regression task, making it more challenging and less explored. Finally, we review relevant multi-scale architectures, which have been widely used in spatiotemporal modeling.

\noindent\textbf{TTC forecasting.} Initial TTC forecasting techniques~\cite{Suzuki2018AnticipatingTA} have depended on agent localization achieved through object tracking and trajectory prediction. With the advent of spatiotemporal modeling advancements, 3D convolutional and two-stream networks have enabled direct extraction of spatial and temporal features. Classic two-stream approaches leverage CNNs for spatial modeling alongside RNNs for temporal dynamics~\cite{anjum2023spatio,tran2015learning}. More recently, attention-based mechanisms have emerged, offering an alternative to conventional convolutions for improved efficiency. Notably, ~\cite{desai2025cyclecrash} introduced a non-stationary attention framework that separates stationary (static) and non-stationary (dynamic) components to refine TTC forecasting.  Despite advancements, TTC forecasting remains challenging, requiring models to capture both short- and long-term spatial-temporal dependencies. While CNNs offer multi-scale feature extraction, the convolutional filters have a limited receptive field. To address this, we propose a hierarchical multi-scale solution for robust TTC forecasting.

\noindent\textbf{Collision anticipation and detection.} Existing works in this area are often classified into Traffic Accident Detection (Vision TAD) and Traffic Accident Anticipation (Vision TAA) tasks. Early Vision TAD methods~\cite{9122456,You2020TrafficAB,Srinivasan2020ANA,Luo2023ASF} utilize frame-level feature analysis to distinguish accident frames from normal ones, often leveraging self-supervised learning or frame-reconstruction techniques~\cite{Singh2019DeepSR,Nguyen2020AnomalyDI,Luo2021FutureFP}. Although recent Vision TAD approaches~\cite{Wang2023DeepAccidentAM,Bajgoti2023SwinAnomalyRV} have integrated pre-trained transformer-based models like Video Swin Transformers~\cite{liu2022video}. However, these methods typically fail to offer proactive collision warnings, which are important for collision prevention.

Conversely, Vision TAA methods have gained prominence for their preventive potential. Traditional TAA approaches~\cite{Taccari2018ClassificationOC,Wu2023PredictingCA,KumaranSanthosh2021VehicularTC,Chakraborty2018FreewayTI,Zeng2017AgentCentricRA} have relied on a combination of object detection, tracking, and trajectory-based accident anticipation, using detectors like YOLO~\cite{yolov3}. For example, \cite{Yi2023ImprovedDS} introduced a Dynamic Spatial Attention module tailored for early accident anticipation. Building on this, \cite{Thakur_2024_WACV} developed a framework incorporating graph convolution and graph attention to enhance prediction capabilities. However, these methods generally lack the capability to predict the exact time of impact, a critical factor for offering actionable insights and guiding effective interventions in imminent collision scenarios.

\noindent \textbf{Multi-scale architectures.}
Multi-scale architectures have already demonstrated notable success in video anomaly detection in a traffic setting~\cite{zhang2024multi,Wang2020RobustUV,Zhong2022BidirectionalSF}. For example, \cite{Wang2020RobustUV} uses features from different image resolutions, and \cite{Zhong2022BidirectionalSF} applies feature pyramids to improve anomaly detection. However, these methods focus on the spatial aspect; they do not capture temporal dynamics, which are crucial for detecting motion and appearance changes over time in traffic scenarios. \cite{zhang2024multi} introduced a multi-scaled spatiotemporal representation learning method that leverages video continuity to design three proxy tasks to perform feature learning at both short- and long-range dependencies, i.e., continuity judgment, discontinuity localization, and missing frame estimation.
Recently, a hierarchical multi-scale Vision Transformer (MViT)~\cite{fan2021multiscale} has been proposed, which can not only handle the global-range bottleneck of CNNs but also learn multi-scale representations over four scales.
It has achieved great benefits on large-scale vision data, although, as transformers lack inductive biases such as variance~\cite{dosovitskiy2021an}, they need data on a larger scale. 
To introduce this inductive bias, we employ a strong pretext task and hierarchically structure the spatial dimension of our transformer, drawing inspiration from~\cite{ryali2023hiera,li2022mvitv2}. This enables CollideNet, our proposed hierarchical multi-scale spatiotemporal method, to achieve robust TTC forecasting.

\label{sec:proposed_method}
\begin{figure*}[ht]
  \centering
    \centering
    \includegraphics[width=0.9\linewidth]{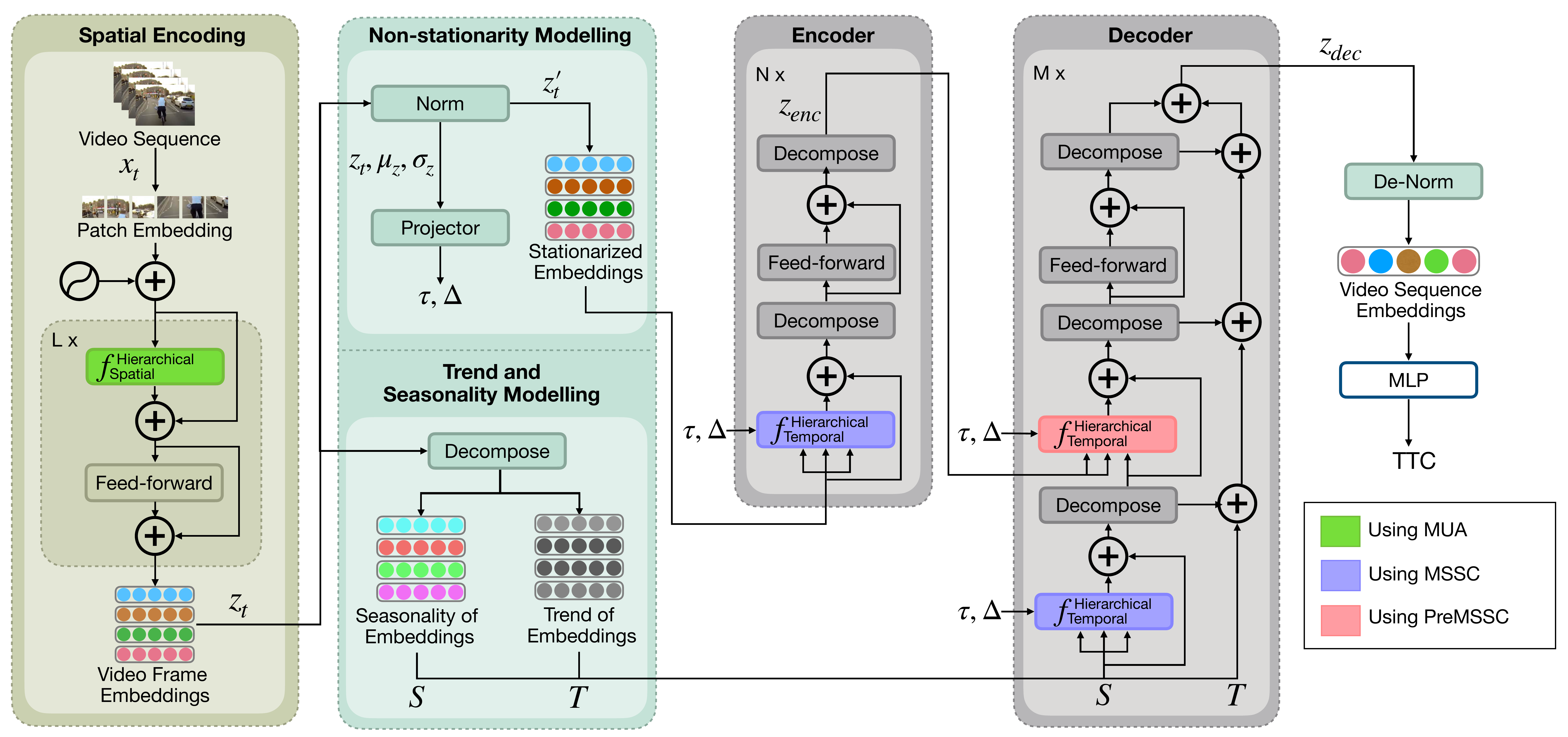}
    \caption{The architecture of the proposed method, CollideNet is presented. First, the hierarchical spatial encoder uses MUA-based attention in  $f_\text{Spatial}^\text{Hierarchical}$ to obtain spatial encodings of the video frames. Next, the video frame embeddings $z_t$ are decomposed into trend $T$ and seasonality $S$, followed by normalization to remove non-stationary effects based on mean $\mu_{z}$ and variance $\sigma_{z}$. This process yields the stationarized time-series $z_t'$. 
    The stationarized time-series is then processed by a multi-scale hierarchical temporal encoder $f_\text{Temporal}^\text{Hierarchical}$, which uses MSSC and PreMSSC.
    Non-stationary information is then reintroduced into the transformer blocks via rescaling factors $\tau$ and $\Delta$, determined by the projector based on the previous normalization layer. Finally, the output is de-normalized using the original statistical parameters.
    }
  \label{fig:method_arch_diag}
\end{figure*}

\section{Method}

Our network architecture employs a two-stream hierarchical multi-scale transformer-based design that captures diverse spatial and temporal scales, integrating global and local spatial features with both long- and short-term temporal dependencies to enable robust TTC forecasting. The overall architecture is shown in Figure~\ref{fig:method_arch_diag}.

\noindent \textbf{Spatial modeling.} Our goal is to efficiently capture both global and local spatial information. To achieve this, we adopt a hierarchical multi-stage approach inspired by prior work
~\cite{li2022mvitv2,ryali2023hiera}, where early stages capture high-resolution local details, 
and later stages extract global features by progressively reducing the number of patches while increasing feature dimensionality at lower resolutions.
This design balances local structure preservation and global context modeling, addressing the inefficiencies of 
conventional ViTs, which typically maintain a constant resolution and patch token dimension. 
Architectural modifications are often added to ViTs, which 
introduce spatial biases to compensate for the lack of inductive bias that ViTs exhibit. 
Instead, here we leverage Masked Autoencoders (MAE)~\cite{9879206} on a strong pretext task, following~\cite{ryali2023hiera}.

The video sequence $x_t$ is first patchified and then embedded with absolute positional encoding, which adds fixed position vectors to image patches to encode their spatial locations in the sequence. We use absolute positional encoding due to its lower computational complexity compared to relative positional encoding, relying on fixed embedded frames extracted over $L$ spatial scales. 
Each scale consists of a hierarchical function $f_\text{Spatial}^\text{Hierarchical}$, which contains transformer blocks with identical resolution across the spatial and patch token dimensions and captures hierarchical representations of spatial information at each stage along with a feed-forward layer. 

Hierarchical transformer architectures require pooling of the query tokens ($Q$) at each stage to progressively reduce dimensionality, while pooling of keys ($K$) and value ($V$) tokens is optional and primarily reduces the size of the attention matrix for computational efficiency~\cite{ryali2023hiera}. For the spatial stream, the Masked Unit Attention (MUA) in $f_\text{Spatial}^\text{Hierarchical}$ uses local attention instead of pooling attention of the original mViTv2 architecture in the initial stages, while global attention is used for the later stages.
Global attention is applied in later stages, as it is more computationally expensive, so switching to local attention early reduces overall computation overhead. The final stage produces the video frame embeddings $z_t$.

\noindent \textbf{Temporal modeling.}
Time series at different scales inherently exhibit distinct properties, with finer scales primarily capturing short-range detailed patterns and coarser scales highlighting long-range variations~\cite{Mozer1991InductionOM}.
Multi-scale architectures have proven to be effective in capturing these intricacies of the temporal dimension, while mitigating the limitations of standard transformers in modeling long-range dependencies~\cite{himtm,Wang2024TimeMixerDM,shabani2023scaleformer,du2023preformer}. 
Temporal disentanglement is a common technique that breaks down a time-series into multiple components, each representing a distinct underlying pattern that is easier to model~\cite{liu2022non,taylor2018forecasting,oreshkin2019n,wu2021autoformer}. 
Low-rank decomposition such as series decomposition~\cite{rb1990stl,Anderson1976TimeSeries2E} has been previously used to separate background (long-range) and foreground (short-ranged) information, improving long-range and short-range context modeling and overcoming issues like partial visibility or occlusions in traffic-related tasks~\cite{6843945,7406471,8265430,Grosek2014DynamicMD}. However, prior work often uses this separation technique in the spatial dimension and leaves the temporal dynamics underexplored. Additionally, this approach can be computationally expensive for high-resolution spatial data. Consequently, unlike prior works, we use series decomposition on video frame embeddings $z_t$ in the temporal dimension to disentangle seasonality (repetitive predictive patterns) $S$ containing the foreground or short-range information and trend (long-term direction) $T$ containing the background or long-range information. This distinction significantly reduces computational complexity. 

Moreover, transformers struggle with forecasting when a real-world time-series is non-stationary, meaning it has changing statistical properties over time~\cite{liu2022non}. While prior work often attenuates the non-stationarity of the time-series to improve predictability, this comes at the cost of neglecting the inherent properties of real-world time-series. Following \cite{liu2022non}, we disentangle the stationary and non-stationary components of the video frame embeddings $z_t$. We use normalization to attenuate the non-stationarity caused by mean and variance, which yields the stationarized embeddings $z_t'$. In order to re-incorporate non-stationary information, we use a linear projector which uses the statistical properties (mean $\mu_z$ and variance $\sigma_z$) of the original video frame embeddings $z_t$ to yield rescaling factors $\tau$ and $\Delta$ which will be later used to re-scale the calculated attention score in the attention mechanism.

\begin{figure}
    \centering
    \begin{subfigure}{0.44\linewidth}
      \centering
      \includegraphics[width=\linewidth]{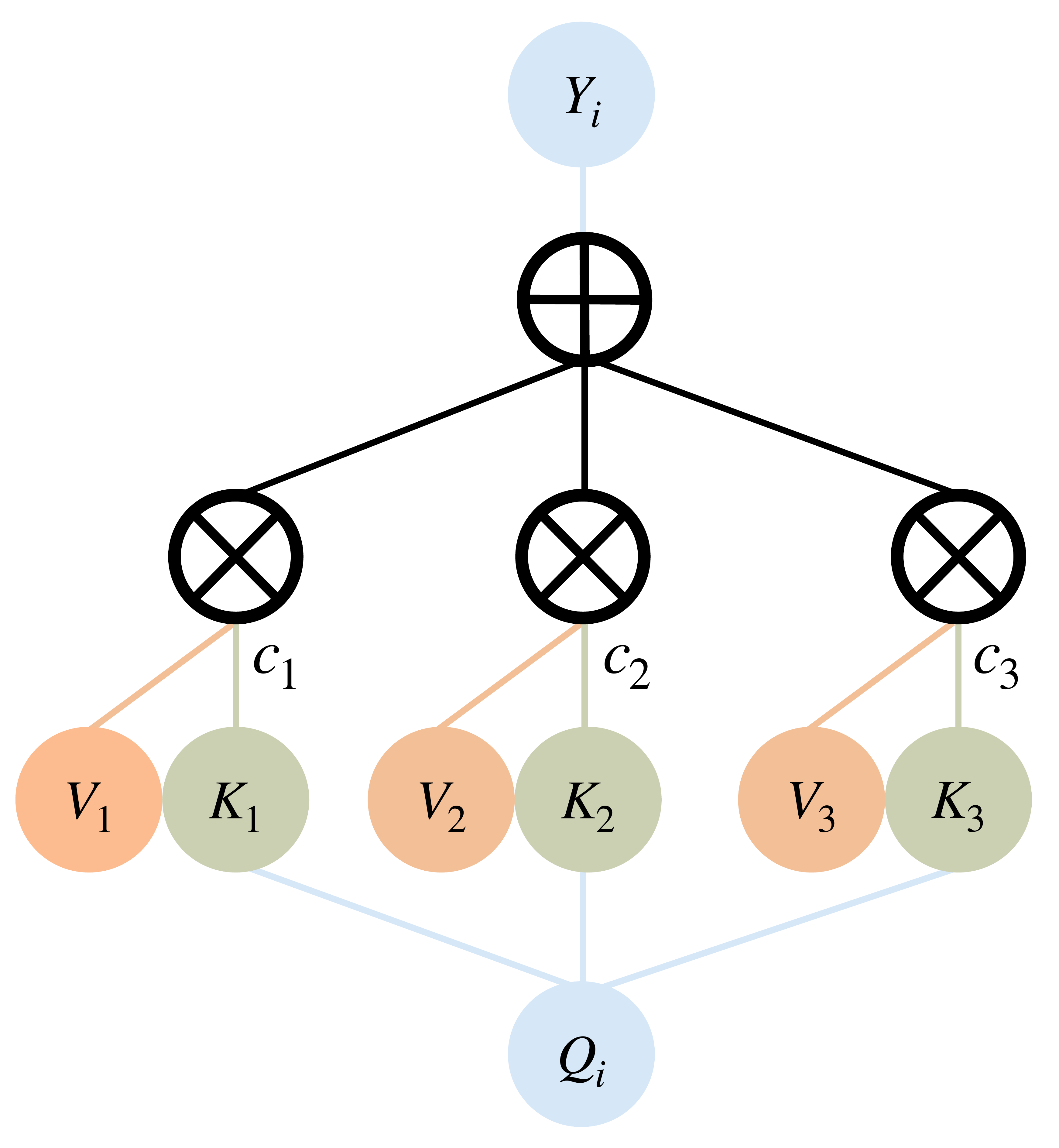}
      \caption{Segment correlation without the predictive paradigm.}
      \label{fig:norm_sc}
    \end{subfigure}
    \hfill
    \begin{subfigure}{0.44\linewidth}
      \centering
      \includegraphics[width=\linewidth]{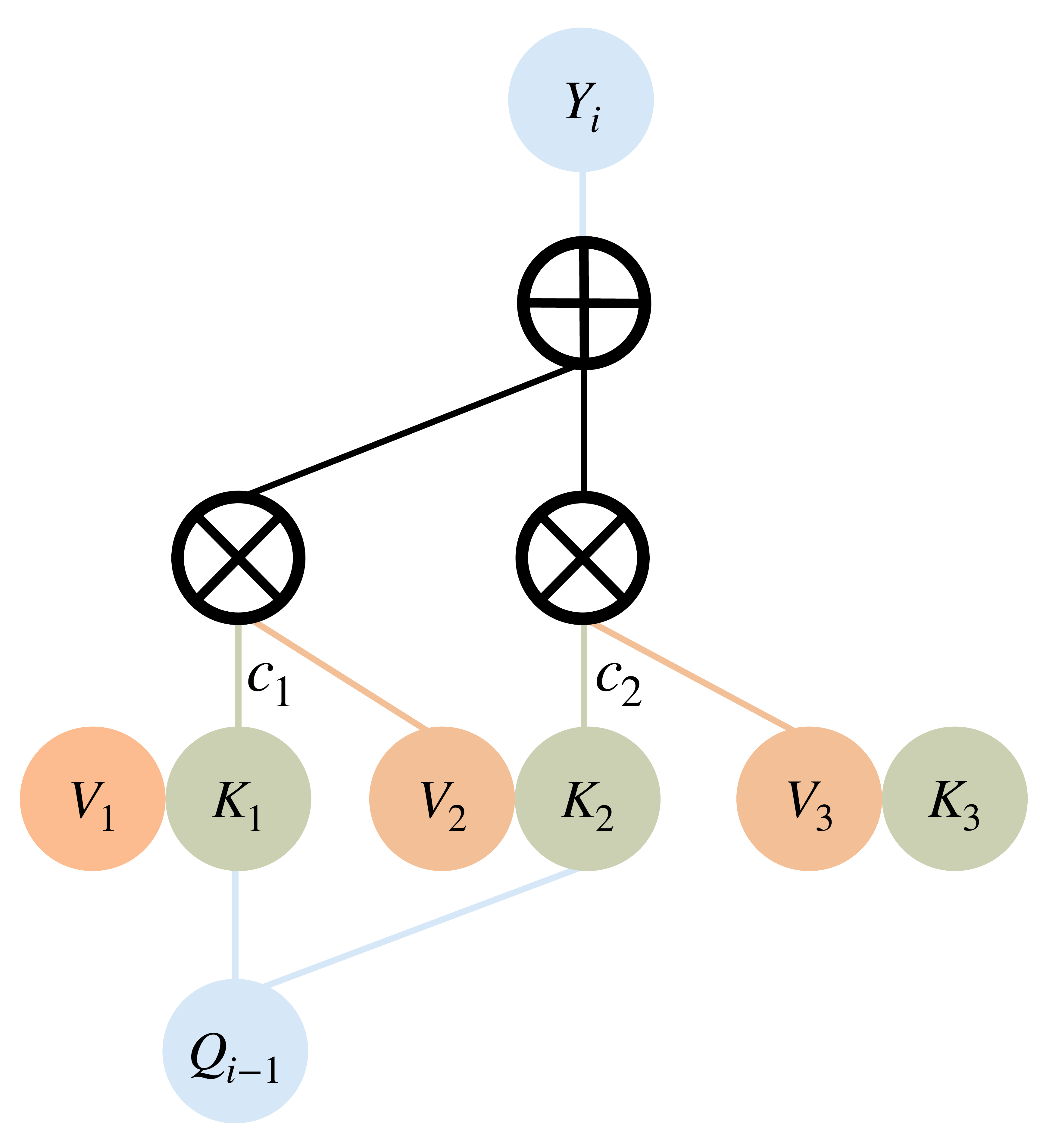}
      \caption{Segment correlation with the predictive paradigm.}
      \label{fig:pred_sc}
    \end{subfigure}
    \caption{The correlation score $c_{j}$
  quantifies the relationship between $Q+{i}$
  and $K_{j}$. 
  Output $Y_{i}$ at the current segment $i$ is determined by using the similarity between $Q_{i-1}$ from the previous segment $i-1$ and the key $K_{j}$ as the weight for the value $V_{j+1}$ in the aggregation process.}
    \label{fig:Segment_correlation}
\end{figure}

For the hierarchical multi-scale temporal encoding, we use a standard Encoder-Decoder framework similar to the original Transformer~\cite{vaswani2017attention} as shown in Figure \ref{fig:method_arch_diag}. The encoder extracts information from historical observations of the stationarized embeddings $z_t'$, while the decoder outputs a unified video sequence embedding. The standard point-wise self-attention mechanism calculates the similarity of each element pair, increasing the computational and space complexity quadratically with the length of the time-series. 
Methods such as \cite{du2023preformer,wu2021autoformer} use a type of sparse attention called segment-wise correlation, which computes similarities between segments of the sequence rather than individual elements (See Figure ~\ref{fig:norm_sc}). Segment-wise correlation also preserves local temporal structure, providing a trade-off between local feature fidelity and computational efficiency.~\cite{du2023preformer}. Here, the length of the series segment determines the resolution. A large value captures long-range dependencies, while a small value captures short-range features. The Multi-Scale Segment-wise Correlation (MSSC) block applies segment-wise correlation at multiple scales, progressively doubling the segment length to capture both short- and long-range dependencies.
The Encoder block consists of $N$ identical layers, each comprising a hierarchical temporal transformer $f_\text{Temporal}^\text{Hierarchical}$ with MSSC and a feed-forward layer, 
followed by a series decomposition module that disentangles the trend and seasonality of the embeddings. 
Following~\cite{du2023preformer,wu2021autoformer}, the encoder discards the trend component to focus solely on modeling seasonal patterns, producing the encoder output $z_{enc}$, which contains the past seasonal information and serves as cross-information to help the decoder refine prediction results. 

The decoder contains $M$ identical layers and receives as input the encoder output $z_{enc}$ along with the $T$ and $S$ components extracted prior to the normalization of the original time series. 
The decoder adopts a structure similar to that of the encoder. The decomposition module in the decoder extracts the trend component after each layer, which is then added to the seasonality part to obtain the decoder output $z_{dec}$.
For forecasting, the query for a predicted time segment corresponds to its previous segment. Following the intuition that influential past segments should weigh more heavily on their future predictions~\cite{du2023preformer}, the decoder incorporates the Predictive MSSC (PreMSSC) block as shown in  Figure~\ref{fig:method_arch_diag}, which is an extension of the MSSC block specialized for forecasting.
This block incorporates cross-attention, introducing a segment lag between the keys and values derived from the encoder output $z_{enc}$. The resulting attention-weighted scores are then shifted back to adjust the initial shift, as shown in Figure~\ref{fig:pred_sc}.
Finally, the De-Normalization module rescales the decoder output $z_{dec}$ using the previously computed mean and variance, producing the final video sequence embedding for the TTC forecasting task.

\section{Experiments}


\noindent \textbf{Dataset and evaluation metrics.}
We use three publicly available datasets, namely \textbf{DAD} \cite{chan2016anticipating}, \textbf{CCD} \cite{BaoMM2020}, and \textbf{DoTA} \cite{yao2022dota}. These datasets are sourced from collision videos on YouTube and come with annotations indicating the time-of-collision, specifying the exact moment a collision begins in each video clip. Following \cite{desai2025cyclecrash}, we upsample the videos of the three datasets to 30 fps and a uniform resolution of 1280$\times$720 pixels. Next, we segment each video into 1-second clips, each assigned a TTC value based on the clip’s start time and the original video’s time-of-collision label. The DAD dataset consists of 1,130 collision videos with 5,540 post-processed clips, CCD contains 1,500 videos with 11,126 clips, and DoTA includes 4,677 videos with 28,455 clips. 
We maintain the provided dataset splits for training and evaluation. Consistent with prior work \cite{desai2025cyclecrash,anjum2023learning,anjum2023spatio}, we use Mean Squared Error (MSE) as the evaluation metric for model performance across all datasets.

\begin{table}[t]
    \centering
\setlength{\tabcolsep}{3pt}{%
    \begin{tabular}{ l l c c c}
        \hline 
        \textbf{Methods} & \textbf{Backbone} & \textbf{CCD} & \textbf{DoTA} & \textbf{DAD} \\ \hline 
        CNN-RNN~\cite{anjum2023learning} & EfficientNet & 0.93 & 1.79 & 0.75 \\ 
        Li3D~\cite{anjum2023learning} & 3D ConvNet & 1.42 & 2.45 & 0.90 \\
        X3D~\cite{feichtenhofer2020x3d} & 3D ConvNet & 0.85 & 2.21 & 0.85 \\
        C3D~\cite{tran2015learning} & 3D ConvNet & 1.98 & 3.59 & 1.75 \\ 
        ResNet50 3D~\cite{hara2017learning} & 3D ConvNet & 0.87 & 2.11 & 0.88 \\ 
        VGG-16~\cite{manglik2019forecasting} & VGG-16 & 0.54 & \underline{1.78} & 0.74 \\  
        ViViT~\cite{arnab2021vivit} & Vision Transformer & 0.82 & 2.13 & 1.02 \\ 
        TimeSformer~\cite{bertasius2021space} & Vision Transformer & 0.63 & 1.98 & 0.79 \\ 
        Video Swin Trans.~\cite{liu2022video} & Swin Transformer & 0.62 & 2.03 & 0.79 \\ 
        HyCT~\cite{anjum2023spatio} & CNN-Transformer & 0.78 & 2.13 & 0.91 \\ 
        Video-FocalNet~\cite{Wasim_2023_ICCV} & Focal Mod. Net & 0.58 & 2.00 & \underline{0.73} \\ 
        VidNeXt~\cite{desai2025cyclecrash} & ConvNeXt & \underline{0.53} & 1.89 & 0.77\\ 
        CollideNet (Ours) & Multi-scale ViT & \textbf{0.37} & \textbf{1.75} & \textbf{0.71} \\ \hline
    \end{tabular}
    }
    \caption{Performance comparison across CCD, DoTA, and DAD datasets. Results are in MSE, where lower is better.}
    \label{tab:methods_comparison}
\end{table}

\noindent \textbf{Baselines.}
We compare our proposed method with previous TTC forecasting methods, such as CNN-RNN~\cite{tran2015learning}, C3D~\cite{tran2015learning}, VGG-16~\cite{manglik2019forecasting}, Li3D~\cite{anjum2023learning}, HyCT~\cite{anjum2023spatio}, 
and VidNeXt~\cite{desai2025cyclecrash}. We also include models widely adopted in the community for video representation learning. In particular, we use convolution-based video learners 
ResNet50 3D~\cite{hara2017learning} and X3D~\cite{feichtenhofer2020x3d} (we use the X3D-M version). 
Additionally, we use transformer-based methods ViViT~\cite{arnab2021vivit}, TimeSformer~\cite{bertasius2021space}, Video Swin Transformer~\cite{liu2022video}, and Video-FocalNet~\cite{Wasim_2023_ICCV}. We specifically include Video-FocalNet~\cite{Wasim_2023_ICCV}, which uses focal modulations that are similar to multi-scale hierarchical architecture to model global and spatial context similar to the proposed method, as well as Video Swin Transformer~\cite{liu2022video}, which uses a spatiotemporal stream for hierarchical transformer in contrast to our proposed two-stream architecture. All these models are converted to TTC baselines by removing the head layer and using the final video sequence embedding, which is connected to a regression head consisting of two MLP layers, each followed by ReLU activation and a dropout layer.

\noindent \textbf{Implementation details.}
We use a pair of NVIDIA A-100 GPUs for training. We use an initial learning rate of $1e^{-6}$, which dynamically decays when the monitored metric stops improving, with a patience of 5 epochs. To prevent overfitting, we apply early stopping with a patience of 15 epochs. Training is conducted for 50 epochs with a batch size of 16, using MSE loss. We employ a step size of 4 and use half-bit mixed precision training to reduce computational overhead. Following  \cite{desai2025cyclecrash}, we handle the class imbalance in the TTC forecasting task by upsampling video clips through duplication and augmenting brightness, contrast, saturation, and hue. All video frames are resized to a resolution of 224$\times$224. 
We use the available implementations of the baseline models in PyTorch.

\begin{table}[t]
    \centering
\setlength{\tabcolsep}{3pt}{%
    \begin{tabular}{ l c c c}
        \hline 
        \textbf{Model}  & \textbf{Params (M)} & \textbf{Latency (ms)} & \textbf{Throughput (samples/sec)}\\ \hline 
\textcolor[RGB]{150,150,150}{CNN-RNN}~\cite{anjum2023learning} & \textcolor[RGB]{150,150,150}{25.20} & 22.09 & 45.26\\
\textcolor[RGB]{150,150,150}{Li3D}~\cite{anjum2023learning} & \textcolor[RGB]{150,150,150}{0.98} & 1.77 & 563.73\\ 
\textcolor[RGB]{150,150,150}{X3D}~\cite{feichtenhofer2020x3d} & \textcolor[RGB]{150,150,150}{3.10} & 15.40 & 64.93\\ 
\textcolor[RGB]{150,150,150}{C3D}~\cite{tran2015learning} & \textcolor[RGB]{150,150,150}{80.00} & 11.08 & 90.26\\ 
\textcolor[RGB]{150,150,150}{ResNet50 3D}~\cite{hara2017learning} & \textcolor[RGB]{150,150,150}{31.80} & 52.10 & 19.19\\ 
\textcolor[RGB]{150,150,150}{VGG16~\cite{manglik2019forecasting}} & \textcolor[RGB]{150,150,150}{15.10} & 30.95 & 32.31\\  
ViViT~\cite{arnab2021vivit} & 114.80 & 43.99 & 22.73\\ 
TimeSformer~\cite{bertasius2021space} & 121.31 & 11.26 & 8.99\\ 
Video Swin Trans.~\cite{liu2022video} & 87.70 & 70.77 & 14.13\\ 
HyCT~\cite{anjum2023spatio} & 23.60 & 198.04 & 5.05\\ 
Video-FocalNet~\cite{Wasim_2023_ICCV} & 156.98 & 138.11 & 7.24\\ 
VidNeXt~\cite{desai2025cyclecrash} & 125.09 & 83.36 & 12.00\\ 
        CollideNet (Ours) & 51.86 & 125.97 & 7.94\\ \hline
    \end{tabular}
    }
    \caption{Comparison of model complexity in Parameters. Convolutional models, which are generally light-weight, are shown in \textcolor[RGB]{150,150,150}{grey}.}
    \label{tab:models_param_comparison}
\end{table}

\section{Results}

\subsection{Performance}

\noindent \textbf{Within-dataset evaluation.}
We compare our proposed method with state-of-the-art methods and widely used spatiotemporal models as baselines for TTC forecasting tasks, measured in MSE.
As shown in Table \ref{tab:methods_comparison}, our proposed method outperforms all baselines across all datasets, achieving the lowest MSE scores of 0.37 on CCD, 1.75 on DoTA, and 0.71 on DAD, respectively. Notably, we observe a significant margin between CollideNet and the second-best method on the CCD dataset, with an improvement of 30\% in MSE.
These results clearly demonstrate the superiority of hierarchical multi-scale modeling, as implemented in CollideNet. Transformer-based approaches, such as VidNeXt~\cite{desai2025cyclecrash} and Video-FocalNet~\cite{Wasim_2023_ICCV}, which yield the second-best performance on the CCD and DAD datasets, respectively, fail to match the consistency of CollideNet.

\noindent \textbf{Model complexity.} Table \ref{tab:models_param_comparison} compares the model complexity in terms of the number of parameters (in millions) for various state-of-the-art models.
CollideNet achieves a reasonable parameter count, which is notably lower compared to transformer-based models like ViViT, TimeSformer and Video-FocalNet, while still delivering state-of-the-art performance. 
This demonstrates that CollideNet's design reduces parameter size while delivering robust predictive capability.

\begin{table}[t]
    \centering
    \setlength{\tabcolsep}{5pt}{%
    \begin{tabular}{l l l c}
        \hline
        \textbf{Method} & \textbf{Training} & \textbf{Eval} & \textbf{MSE} \\ \hline 
        Video-FocalNet & CCD & DoTA & 2.047 \\
        Video-FocalNet & CCD & DAD & 0.787 \\
        Video-FocalNet & DoTA & CCD & 0.563 \\
        Video-FocalNet & DoTA & DAD & 0.760 \\
        Video-FocalNet & DAD & CCD & 0.603 \\
        Video-FocalNet & DAD & DoTA & 1.984 \\ \hline
        VidNeXt & CCD & DoTA & 1.988 \\
        VidNeXt & CCD & DAD & 0.817 \\
        VidNeXt & DoTA & CCD & 0.552 \\
        VidNeXt & DoTA & DAD & 0.813 \\
        VidNeXt & DAD & CCD & 0.604 \\
        VidNeXt & DAD & DoTA & 2.020 \\ \hline
        CollideNet & CCD & DoTA & 1.711 \\
        CollideNet & CCD & DAD & 0.731 \\
        CollideNet & DoTA & CCD & 0.382 \\
        CollideNet & DoTA & DAD & 0.727 \\
        CollideNet & DAD & CCD & 0.439 \\
        CollideNet & DAD & DoTA & 1.761 \\ \hline
    \end{tabular}
    }
    \caption{Cross-dataset evaluation.}
    \label{tab:cross_dataset_evaluation}
\end{table}

\noindent \textbf{Cross-dataset evaluation.}
Table \ref{tab:cross_dataset_evaluation} presents a cross-dataset evaluation.
The results reveal significant MSE variations across datasets, highlighting domain shift challenges in video-based TTC tasks. We observe that CollideNet demonstrates superior generalization, achieving the lowest MSE in most transfer scenarios, including an MSE of 1.711 on CCD-to-DoTA, outperforming its performance on DoTA. This indicates CollideNet’s ability to extract more transferable features that generalize effectively across domains by consistently mitigating domain shift degradation compared to other methods. While Video-FocalNet and VidNeXt show competitive performance, VidNeXt slightly outperforms Video-FocalNet on CCD and DoTA, while both methods struggle with consistency across datasets.

\section{Conclusion}
In this work, we tackle the critical task of TTC forecasting, a key challenge in collision prevention that requires precise temporal predictions for timely intervention. Recognizing the inherently multi-scale nature of video data, which encapsulates a diverse range from long- to short-range dependencies, we propose CollideNet, a novel two-stream hierarchical method designed to address limitations in current state-of-the-art approaches and to effectively capture multi-scale spatiotemporal features while disentangling and modeling trend, seasonality, and non-stationarity components.
We validate the effectiveness of CollideNet through comprehensive experiments on three benchmark datasets, outperforming all prior methods. 
Additionally, cross-dataset evaluations underscore CollideNet's superior generalization capabilities, while visualizations reveal the benefits of trend and seasonality disentanglement.

\section*{Acknowledgments}
We thank Geotab Inc.,
the City of Kingston, and NSERC for supporting this work, as well as Debaditya Shome and Pritam Sarkar for their contributions.

{   
    \bibliographystyle{splncs04}
    \bibliography{main}

@misc{nhtsa2024,
  author       = {{National Highway Traffic Safety Administration (NHTSA)}},
  title        = {{Automated Vehicles for Safety}},
  year         = 2024,
  url          = {https://www.nhtsa.gov/vehicle-safety/automated-vehicles-safety},
  urldate      = {2024-12-24},
  note         = {{Accessed: 2024-12-24}}
}

@article{teoh2021effectiveness,
  title = {{Effectiveness of Front Crash Prevention Systems in Reducing Large Truck Real-World Crash Rates}},
  author = {Teoh, Eric R.},
  journal = {{Traffic Injury Prevention}},
  year = {2021},
  month = {March},
  url = {https://www.iihs.org/topics/bibliography/ref/2211},
  note = {{Insurance Institute for Highway Safety, Highway Loss Data Institute, ID: 2211}}
}

@misc{ntsb2001,
  author       = {{National Transportation Safety Board}},
  title        = {{Special Investigation Report: Highway Vehicle and Infrastructure-based Technology for the Prevention of Rear-end Collisions}},
  institution  = {National Transportation Safety Board},
  year         = {2001},
  type         = {Safety Recommendation Report},
  number       = {SIR-01/01},
  url          = {http://www.ntsb.gov/Publictn/2001/SIR0101.pdf}
}

@article{janssen1994ttc,
  author       = {Janssen, W. and Thomas, M.},
  title        = {{Time-to-Collision and Collision Avoidance Systems}},
  journal      = {{International Co-operation on Theories and Concepts in Traffic Safety Workshop}},
  year         = {1994},
}

@inproceedings{zhang2024multi,
  title={{Multi-Scale Video Anomaly Detection by Multi-Grained Spatio-Temporal Representation Learning}},
  author={Zhang, Menghao and Wang, Jingyu and Qi, Qi and Sun, Haifeng and Zhuang, Zirui and Ren, Pengfei and Ma, Ruilong and Liao, Jianxin},
  booktitle={{IEEE/CVF Conference on Computer Vision and Pattern Recognition}},
  pages={17385--17394},
  year={2024}
}

@article{Qiu2019LearningSR,
  title={{Learning Spatio-Temporal Representation With Local and Global Diffusion}},
  author={Zhaofan Qiu and Ting Yao and Chong-Wah Ngo and Xinmei Tian and Tao Mei},
  journal={{IEEE/CVF Conference on Computer Vision and Pattern Recognition }},
  year={2019},
  pages={12048-12057},
}

@inproceedings{manglik2019forecasting,
  title={{Forecasting Time-to-Collision from Monocular Video: Feasibility, Dataset, and Challenges}},
  author={Manglik, Aashi and Weng, Xinshuo and Ohn-Bar, Eshed and Kitanil, Kris M},
  booktitle={{IEEE/RSJ International Conference on Intelligent Robots and Systems}},
  pages={8081--8088},
  year={2019},
}

@inproceedings{tran2015learning,
  title={{Learning Spatiotemporal Features with 3d Convolutional Networks}},
  author={Tran, Du and Bourdev, Lubomir and Fergus, Rob and Torresani, Lorenzo and Paluri, Manohar},
  booktitle={{IEEE International Conference on Computer Vision}},
  pages={4489--4497},
  year={2015}
}

@inproceedings{hara2017learning,
  title={{Learning Spatio-temporal Features with 3d Residual Networks for Action Recognition}},
  author={Hara, Kensho and Kataoka, Hirokatsu and Satoh, Yutaka},
  booktitle={{IEEE International Conference on Computer Vision Workshops}},
  pages={3154--3160},
  year={2017}
}

@inproceedings{bertasius2021space,
  title={{Is Space-time Attention All You Need for Video Understanding?}},
  author={Bertasius, Gedas and Wang, Heng and Torresani, Lorenzo},
  booktitle={{International Conference on Machine Learning}},
  vol=2,
  vol=2,
  year={2021}
}

@inproceedings{liu2022video,
  title={{Video Swin Transformer}},
  author={Liu, Ze and Ning, Jia and Cao, Yue and Wei, Yixuan and Zhang, Zheng and Lin, Stephen and Hu, Han},
  booktitle={{IEEE/CVF Conference on Computer Vision and Pattern Recognition}},
  pages={3202--3211},
  year={2022}
}

@inproceedings{anjum2023learning,
  title={{Learning Spatio-Temporal Features via 3D CNNs to Forecast Time-to-Accident}},
  author={Anjum, Taif and Chirade, Louis and Lin, Beiyu and Narayan, Apurva},
  booktitle={{International Conference on Agents and Artificial Intelligence}},
  pages={532--540},
  year={2023}
}

@inproceedings{desai2025cyclecrash,
  title={Cyclecrash: A dataset of bicycle collision videos for collision prediction and analysis},
  author={Desai, Nishq Poorav and Etemad, Ali and Greenspan, Michael},
  booktitle={2025 IEEE/CVF Winter Conference on Applications of Computer Vision (WACV)},
  pages={6688--6698},
  year={2025},
}

@InProceedings{Wasim_2023_ICCV,
    author    = {Wasim, Syed Talal and Khattak, Muhammad Uzair and Naseer, Muzammal and Khan, Salman and Shah, Mubarak and Khan, Fahad Shahbaz},
    title     = {{Video-FocalNets: Spatio-Temporal Focal Modulation for Video Action Recognition}},
    booktitle = {{IEEE/CVF International Conference on Computer Vision}},
    year      = {2023},
}

@inproceedings{fan2021multiscale,
  title={{Multiscale Vision Transformers}},
  author={Fan, Haoqi and Xiong, Bo and Mangalam, Karttikeya and Li, Yanghao and Yan, Zhicheng and Malik, Jitendra and Feichtenhofer, Christoph},
  booktitle={{IEEE/CVF International Conference on Computer Vision}},
  pages={6824--6835},
  year={2021}
}

@inproceedings{ryali2023hiera,
  title={{Hiera: A Hierarchical Vision Transformer Without the Bells-and-Whistles}},
  author={Ryali, Chaitanya and Hu, Yuan-Ting and Bolya, Daniel and Wei, Chen and Fan, Haoqi and Huang, Po-Yao and Aggarwal, Vaibhav and Chowdhury, Arkabandhu and Poursaeed, Omid and Hoffman, Judy and others},
  booktitle={{International Conference on Machine Learning}},
  pages={29441--29454},
  year={2023},
}

@inproceedings{li2022mvitv2,
  title={{mViTv2: Improved Multiscale Vision Transformers for Classification and Detection}},
  author={Li, Yanghao and Wu, Chao-Yuan and Fan, Haoqi and Mangalam, Karttikeya and Xiong, Bo and Malik, Jitendra and Feichtenhofer, Christoph},
  booktitle={{IEEE/CVF Conference on Computer Vision and Pattern Recognition}},
  pages={4804--4814},
  year={2022}
}

@inproceedings{du2023preformer,
  title={{Preformer: Predictive Transformer with Multi-scale Segment-wise Correlations for Long-term Time Series Forecasting}},
  author={Du, Dazhao and Su, Bing and Wei, Zhewei},
  booktitle={{IEEE International Conference on Acoustics, Speech and Signal Processing }},
  pages={1--5},
  year={2023},
}

@article{taylor2018forecasting,
  title={{Forecasting at Scale}},
  author={Taylor, Sean J and Letham, Benjamin},
  journal={{The American Statistician}},
  volume={72},
  number={1},
  pages={37--45},
  year={2018},
}

@article{oreshkin2019n,
  title={{N-BEATS: Neural Basis Expansion Analysis for Interpretable Time Series Forecasting}},
  author={Oreshkin, Boris N and Carpov, Dmitri and Chapados, Nicolas and Bengio, Yoshua},
  journal={arXiv preprint arXiv:1905.10437},
  year={2019}
}

@inproceedings{Wang2023DeepAccidentAM,
  title={{DeepAccident: A Motion and Accident Prediction Benchmark for V2X Autonomous Driving}},
  author={Tianqi Wang and Sukmin Kim and Wenxuan Ji and Enze Xie and Chongjian Ge and Junsong Chen and Zhenguo Li and Ping Luo},
  booktitle={{Association for the Advancement of Artificial Intelligence Conference on Artificial Intelligence}},
  year={2023},
}

@InProceedings{Thakur_2024_WACV,
    author    = {Thakur, Nupur and Gouripeddi, PrasanthSai and Li, Baoxin},
    title     = {{Graph(Graph): A Nested Graph-Based Framework for Early Accident Anticipation}},
    booktitle = {{IEEE/CVF Winter Conference on Applications of Computer Vision}},
    year      = {2024},
    pages     = {7533-7541}
}

@article{Suzuki2018AnticipatingTA,
  title={{Anticipating Traffic Accidents with Adaptive Loss and Large-Scale Incident DB}},
  author={Tomoyuki Suzuki and Hirokatsu Kataoka and Yoshimitsu Aoki and Yutaka Satoh},
  journal={{IEEE/CVF Conference on Computer Vision and Pattern Recognition}},
  year={2018},
  pages={3521-3529},
}

@ARTICLE{9122456,
  author={Wang, Xin and Liu, Jing and Qiu, Tie and Mu, Chaoxu and Chen, Chen and Zhou, Pan},
  journal={{IEEE Transactions on Vehicular Technology}}, 
  title={{A Real-Time Collision Prediction Mechanism With Deep Learning for Intelligent Transportation System}}, 
  year={2020},
  volume={69},
  pages={9497-9508}}

@article{yolov3,
  title={{YOLOv3: An Incremental Improvement}},
  author={Redmon, Joseph and Farhadi, Ali},
  journal={{arXiv preprint arXiv:2004.10934}},
  year={2018}
}

@inproceedings{You2020TrafficAB,
  title={{Traffic Accident Benchmark for Causality Recognition}},
  author={Tackgeun You and Bohyung Han},
  booktitle={{European Conference on Computer Vision}},
  year={2020},
}

@article{Srinivasan2020ANA,
  title={{A Novel Approach for Road Accident Detection using DETR Algorithm}},
  author={Aparajith Srinivasan and Anirudh Srikanth and Haresh Indrajit and Venkateswaran Narasimhan},
  journal={{International Conference on Intelligent Data Science Technologies and Applications}},
  year={2020},
  pages={75-80},
}

@article{Singh2019DeepSR,
  title={{Deep Spatio-Temporal Representation for Detection of Road Accidents Using Stacked Autoencoder}},
  author={Dinesh Singh and Chalavadi Krishna Mohan},
  journal={{IEEE Transactions on Intelligent Transportation Systems}},
  year={2019},
  volume={20},
  pages={879-887},
}

@article{Luo2023ASF,
  title={{A Simulation-Based Framework for Urban Traffic Accident Detection}},
  author={Haohan Luo and Feng Wang},
  journal={{IEEE International Conference on Acoustics, Speech and Signal Processing}},
  year={2023},
  pages={1-5},
}

@article{Nguyen2020AnomalyDI,
  title={{Anomaly Detection in Traffic Surveillance Videos with GAN-based Future Frame Prediction}},
  author={Khac-Tuan Nguyen and Dat-Thanh Dinh and Minh N. Do and Minh-Triet Tran},
  journal={{International Conference on Multimedia Retrieval}},
  year={2020},
}

@article{Luo2021FutureFP,
  title={{Future Frame Prediction Network for Video Anomaly Detection}},
  author={Weixin Luo and Wen Liu and Dongze Lian and Shenghua Gao},
  journal={{IEEE Transactions on Pattern Analysis and Machine Intelligence}},
  year={2021},
  volume={44},
  pages={7505-7520},
}

@article{Bajgoti2023SwinAnomalyRV,
  title={{SwinAnomaly: Real-Time Video Anomaly Detection Using Video Swin Transformer and SORT}},
  author={Arpit Bajgoti and Rishik Gupta and Prasanalakshmi B and Rinky Dwivedi and Meena Siwach and Deepak Gupta},
  journal={{IEEE Access}},
  year={2023},
  volume={11},
  pages={111093-111105},
}

@article{KumaranSanthosh2021VehicularTC,
  title={{Vehicular Trajectory Classification and Traffic Anomaly Detection in Videos Using a Hybrid CNN-VAE Architecture}},
  author={Kelathodi Kumaran Santhosh and Debi Prosad Dogra and Partha Pratim Roy and Adway Mitra},
  journal={{IEEE Transactions on Intelligent Transportation Systems}},
  year={2021},
  volume={23},
  pages={11891-11902},
}

@article{Chakraborty2018FreewayTI,
  title={{Freeway Traffic Incident Detection from Cameras: A Semi-Supervised Learning Approach}},
  author={Pranamesh Chakraborty and Anuj Sharma and Chinmay Hegde},
  journal={{International Conference on Intelligent Transportation Systems}},
  year={2018},
  pages={1840-1845},
}

@article{Yi2023ImprovedDS,
  title={{Improved Dynamic Spatial-Temporal Attention Network for Early Anticipation of Traffic Accidents}},
  author={Chao Yi and Ting Huang and Han-Jia Ye and De-chuan Zhan},
  journal={{IEEE International Conference on Multimedia and Expo Workshops}},
  year={2023},
  pages={81-86},
}

@article{Taccari2018ClassificationOC,
  title={{Classification of Crash and Near-Crash Events from Dashcam Videos and Telematics}},
  author={Leonardo Taccari and Francesco Sambo and Luca Bravi and Samuele Salti and Leonardo Sarti and Matteo Simoncini and Alessandro Lori},
  journal={{International Conference on Intelligent Transportation Systems}},
  year={2018},
  pages={2460-2465},
}

@article{Wu2023PredictingCA,
  title={{Predicting Car Accidents with YOLOv7 Object Detection and Object Relationships}},
  author={Ming-Xuan Wu and Chia-Sheng Chang and J. M. Miao and Chia-Yen Lee},
  journal={{IEEE International Conference on Multimedia and Expo Workshops}},
  year={2023},
  pages={87-89},
}

@article{Zeng2017AgentCentricRA,
  title={{Agent-Centric Risk Assessment: Accident Anticipation and Risky Region Localization}},
  author={Kuo-Hao Zeng and Shih-Han Chou and Fu-Hsiang Chan and Juan Carlos Niebles and Min Sun},
  journal={{IEEE Conference on Computer Vision and Pattern Recognition }},
  year={2017},
  pages={1330-1338},
}

@article{rb1990stl,
  title={{STL: A Seasonal-Trend Decomposition Procedure Based on Loess}},
  author={RB, CLEVELAND},
  journal={{Journal Of Statistics}},
  volume={6},
  pages={3--73},
  year={1990}
}

@ARTICLE{6843945,
  author={Wen, Jiajun and Xu, Yong and Tang, Jinhui and Zhan, Yinwei and Lai, Zhihui and Guo, Xiaotang},
  journal={{IEEE Transactions on Circuits and Systems for Video Technology}}, 
  title={{Joint Video Frame Set Division and Low-Rank Decomposition for Background Subtraction}}, 
  year={2014},
  volume={24},
  number={12},
  pages={2034-2048}}

@INPROCEEDINGS{7406471,
  author={Kutz, J. Nathan and Fu, Xing and Brunton, Steve L. and Erichson, N. Benjamin},
  booktitle={{IEEE International Conference on Computer Vision Workshop}}, 
  title={{Multi-resolution Dynamic Mode Decomposition for Foreground/Background Separation and Object Tracking}}, 
  year={2015},
  volume={},
  number={},
  pages={921-929}}

@INPROCEEDINGS{8265430,
  author={Pendergrass, S. and Brunton, S. L. and Kutz, J. N. and Erichson, N. B. and Askham, T.},
  booktitle={{IEEE International Conference on Computer Vision Workshops}}, 
  title={{Dynamic Mode Decomposition for Background Modeling}}, 
  year={2017},
  volume={},
  number={},
  pages={1862-1870}}

@article{Grosek2014DynamicMD,
  title={{Dynamic Mode Decomposition for Real-Time Background/Foreground Separation in Video}},
  author={Jacob Grosek and J. Nathan Kutz},
  journal={{arXiv preprint arXiv:1404.7592}},
  year={2014},
}

@article{Nagar2024SEMASA,
  title={{SEMA: Semantic Attention for Capturing Long-Range Dependencies in Egocentric Lifelogs}},
  author={Pravin Nagar and Ajay Shastry and Jayesh Chaudhari and Chetan Arora},
  journal={{IEEE/CVF Winter Conference on Applications of Computer Vision}},
  year={2024},
  pages={7010-7020},
}

@article{Yoo2023TowardsEG,
  title={{Towards End-to-End Generative Modeling of Long Videos with Memory-Efficient Bidirectional Transformers}},
  author={Jae Hyeon Yoo and Semin Kim and Doyup Lee and Chiheon Kim and Seunghoon Hong},
  journal={{IEEE/CVF Conference on Computer Vision and Pattern Recognition}},
  year={2023},
  pages={22888-22897},
}

@article{Qin2023FactorizationVT,
  title={{Factorization Vision Transformer: Modeling Long Range Dependency with Local Window Cost}},
  author={Haolin Qin and Daquan Zhou and Tingfa Xu and Ziyang Bian and Jianan Li},
  journal={{IEEE Transactions on Neural Networks and Learning Systems}},
  year={2023},
}

@inproceedings{
dosovitskiy2021an,
title={{An Image is Worth 16x16 Words: Transformers for Image Recognition at Scale}},
author={Alexey Dosovitskiy and Lucas Beyer and Alexander Kolesnikov and Dirk Weissenborn and Xiaohua Zhai and Thomas Unterthiner and Mostafa Dehghani and Matthias Minderer and Georg Heigold and Sylvain Gelly and Jakob Uszkoreit and Neil Houlsby},
booktitle={{International Conference on Learning Representations}},
year={2021},
}

@article{Wang2020RobustUV,
  title={{Robust Unsupervised Video Anomaly Detection by Multipath Frame Prediction}},
  author={X. Wang and Zhengping Che and Ke Yang and Bo Jiang and Jian-Bo Tang and Jieping Ye and Jingyu Wang and Q. Qi},
  journal={{IEEE Transactions on Neural Networks and Learning Systems}},
  year={2020},
  volume={33},
  pages={2301-2312},
}

@article{Zhong2022BidirectionalSF,
  title={{Bidirectional Spatio-Temporal Feature Learning With Multiscale Evaluation for Video Anomaly Detection}},
  author={Yuanhong Zhong and Xia Chen and Yongting Hu and Panliang Tang and Fan Ren},
  journal={{IEEE Transactions on Circuits and Systems for Video Technology}},
  year={2022},
  volume={32},
  pages={8285-8296},
}

@inproceedings{feichtenhofer2020x3d,
  title={{X3D: Expanding Architectures for Efficient Video Recognition}},
  author={Feichtenhofer, Christoph},
  booktitle={IEEE/CVF Conference on Computer Vision and Pattern Recognition},
  year={2020}
}

@inproceedings{anjum2023spatio,
  title={{Spatio-temporal Analysis of Dashboard Camera Videos for Time-To-Accident Forecasting}},
  author={Anjum, Taif and Kumar, Daya and Narayan, Apurva},
  booktitle={{IEEE International Joint Conference on Neural Networks}},
  pages={1--8},
  year={2023},
}

@article{liu2022non,
  title={{Non-stationary transformers: Exploring the Stationarity in Time Series Forecasting}},
  author={Liu, Yong and Wu, Haixu and Wang, Jianmin and Long, Mingsheng},
  journal={{Advances in Neural Information Processing Systems}},
  volume={35},
  pages={9881--9893},
  year={2022}
}

@article{vaswani2017attention,
  title={{Attention is All You Need}},
  author={Vaswani, A},
  journal={{Advances in Neural Information Processing Systems}},
  year={2017}
}

@inproceedings{arnab2021vivit,
  title={{ViViT: A video vision transformer}},
  author={Arnab, Anurag and Dehghani, Mostafa and Heigold, Georg and Sun, Chen and Lu{\v{c}}i{\'c}, Mario and Schmid, Cordelia},
  booktitle={{IEEE/CVF International Conference on Computer Vision}},
  pages={6836--6846},
  year={2021}
}

@inproceedings{Wang2023MICNML,
  title={{MICN: Multi-scale Local and Global Context Modeling for Long-term Series Forecasting}},
  author={Huiqiang Wang and Jian Peng and Feihu Huang and Jince Wang and Junhui Chen and Yifei Xiao},
  booktitle={{International Conference on Learning Representations}},
  year={2023},
}

@InProceedings{BaoMM2020,
    author = {Bao, Wentao and Yu, Qi and Kong, Yu},
    title  = {{Uncertainty-based Traffic Accident Anticipation with Spatio-Temporal Relational Learning}},
    booktitle = {ACM Multimedia Conference},
    month  = {May},
    year   = {2020}
}

@article{yao2022dota,
  title={{DoTA: unsupervised detection of traffic anomaly in driving videos}},
  author={Yao, Yu and Wang, Xizi and Xu, Mingze and Pu, Zelin and Wang, Yuchen and Atkins, Ella and Crandall, David},
  journal={IEEE Transactions on Pattern Analysis and Machine Intelligence},
  year={2022},
}

@inproceedings{chan2016anticipating,
    title={{Anticipating Accidents in Dashcam Videos}},
    author={Chan, Fu-Hsiang and Chen, Yu-Ting and Xiang, Yu and Sun, Min},
    booktitle={{Asian Conference on Computer Vision}},
    pages={136--153},
    year={2016},
}

@inproceedings{Mozer1991InductionOM,
  title={{Induction of Multiscale Temporal Structure}},
  author={Michael C. Mozer},
  booktitle={Neural Information Processing Systems},
  year={1991},
}

@INPROCEEDINGS{9879206,
  author={He, Kaiming and Chen, Xinlei and Xie, Saining and Li, Yanghao and Dollár, Piotr and Girshick, Ross},
  booktitle={{IEEE/CVF Conference on Computer Vision and Pattern Recognition}}, 
  title={Masked Autoencoders Are Scalable Vision Learners}, 
  year={2022},
  pages={15979-15988},}

@inproceedings{himtm,
author = {Zhao, Shubao and Jin, Ming and Hou, Zhaoxiang and Yang, Chengyi and Li, Zengxiang and Wen, Qingsong and Wang, Yi},
title = {{HiMTM: Hierarchical Multi-Scale Masked Time Series Modeling with Self-Distillation for Long-Term Forecasting}},
year = {2024},
booktitle = {ACM International Conference on Information and Knowledge Management},
pages = {3352–3362},
numpages = {11},
}

@article{Wang2024TimeMixerDM,
  title={{TimeMixer: Decomposable Multiscale Mixing for Time Series Forecasting}},
  author={Shiyu Wang and Haixu Wu and Xiao Long Shi and Tengge Hu and Huakun Luo and Lintao Ma and James Y. Zhang and Jun Zhou},
  journal={ArXiv},
  year={2024},
  volume={abs/2405.14616},
}

@inproceedings{
shabani2023scaleformer,
title={{Scaleformer: Iterative Multi-scale Refining Transformers for Time Series Forecasting}},
author={Mohammad Amin Shabani and Amir H. Abdi and Lili Meng and Tristan Sylvain},
booktitle={International Conference on Learning Representations },
year={2023},
}

@inproceedings{
wu2021autoformer,
title={{Autoformer: Decomposition Transformers with Auto-Correlation for Long-Term Series Forecasting}},
author={Haixu Wu and Jiehui Xu and Jianmin Wang and Mingsheng Long},
booktitle={{Advances in Neural Information Processing Systems}},
editor={A. Beygelzimer and Y. Dauphin and P. Liang and J. Wortman Vaughan},
year={2021},
}

@article{Anderson1976TimeSeries2E,
  title={{Time-Series.}},
  author={Oliver D. Anderson and M. G. Kendall},
  journal={The Statistician},
  year={1976},
  volume={25},
  pages={308},
}

@article{Kay2017TheKH,
  title={{The Kinetics Human Action Video Dataset}},
  author={Will Kay and Jo{\~a}o Carreira and Karen Simonyan and Brian Zhang and Chloe Hillier and Sudheendra Vijayanarasimhan and Fabio Viola and Tim Green and Trevor Back and Apostol Natsev and Mustafa Suleyman and Andrew Zisserman},
  journal={{arXiv preprint arXiv:1705.06950}},
  year={2017},
}

@inproceedings{Radford2021LearningTV,
  title={{Learning Transferable Visual Models from Natural Language Supervision}},
  author={Alec Radford and Jong Wook Kim and Chris Hallacy and Aditya Ramesh and Gabriel Goh and Sandhini Agarwal and Girish Sastry and Amanda Askell and Pamela Mishkin and Jack Clark and Gretchen Krueger and Ilya Sutskever},
  booktitle={{International Conference on Machine Learning}},
  year={2021}
}

@article{ADFTest,
 author = {Said E. Said and David A. Dickey},
 journal = {Biometrika},
 number = {3},
 pages = {599--607},
 title = {{Testing for Unit Roots in Autoregressive-Moving Average Models of Unknown Order}},
 volume = {71},
 year = {1984}
}

@article{KWIATKOWSKI1992159,
title = {{Testing the Null Hypothesis of Stationarity Against the Alternative of a Unit Root: How Sure Are We That Economic Time Series Have a Unit Root?}},
journal = {{Journal of Econometrics}},
volume = {54},
number = {1},
pages = {159-178},
year = {1992},
author = {Denis Kwiatkowski and Peter C.B. Phillips and Peter Schmidt and Yongcheol Shin},
}

@misc{von-platen-etal-2022-diffusers,
  author = {Patrick von Platen and Suraj Patil and Anton Lozhkov and Pedro Cuenca and Nathan Lambert and Kashif Rasul and Mishig Davaadorj and Dhruv Nair and Sayak Paul and William Berman and Yiyi Xu and Steven Liu and Thomas Wolf},
  title = {{Diffusers: State-of-the-art Diffusion Models}},
  year = {2022},
  journal = {{GitHub repository}},
  howpublished = {\url{https://github.com/huggingface/diffusers}}
}
}
\clearpage
\appendix

\renewcommand{\thefigure}{A\arabic{figure}}
\renewcommand{\thetable}{A\arabic{table}}
\setcounter{figure}{0}
\setcounter{table}{0}

\section*{Appendix}

\subsection*{A.1. Release details}
The code to implement the proposed method is publicly available and can be accessed at \url{https://github.com/DeSinister/CollideNet/}. The initial release comprises:
\begin{itemize}
    \item PyTorch-based implementation for CollideNet.
    \item PyTorch-based implementations for the baselines.
    \item PyTorch-based implementation for the pre-processing
used in the paper, and data loader codes for efficiency
and consistency.
\end{itemize}

\subsection*{A.2 t-SNE visualization of the datasets}
\begin{figure}[h]
      \centering
      \includegraphics[width=0.9\linewidth]{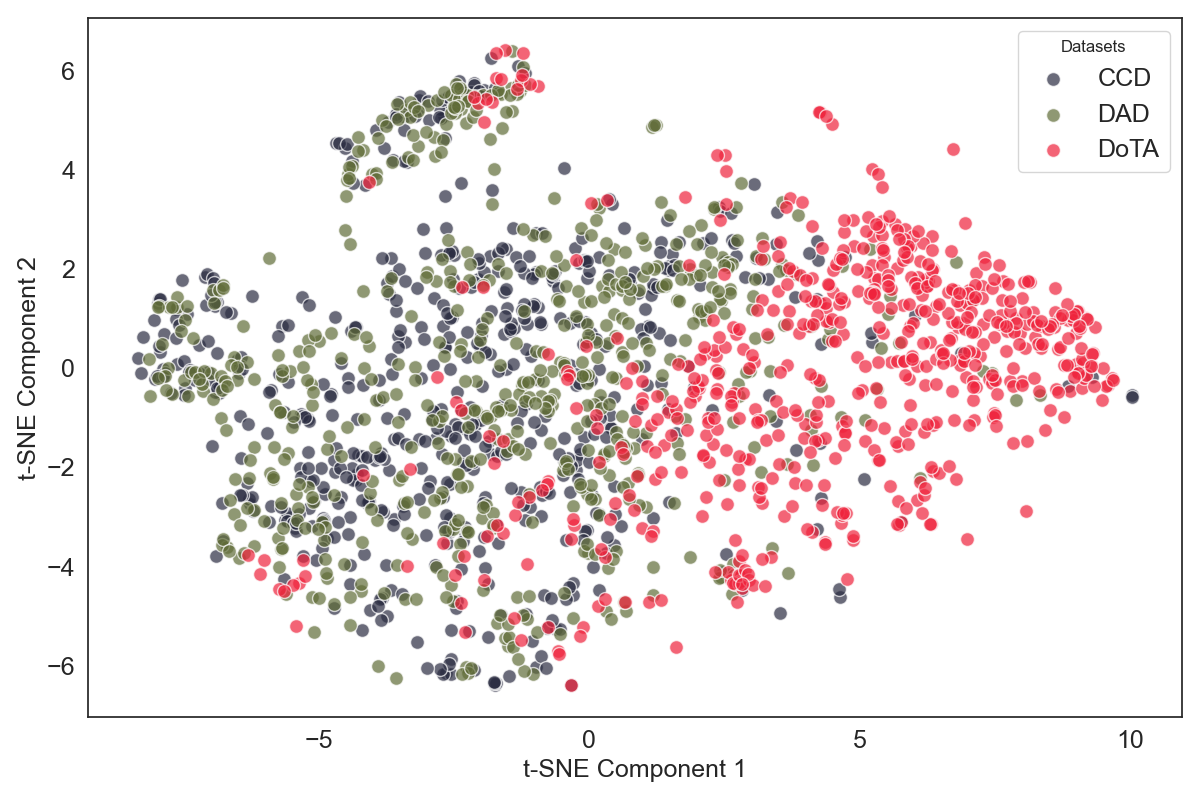}
      \caption{tSNE representations from the CCD, DoTA, and DAD datasets.}
      \label{fig:tsne}
\end{figure}
To explore the nature of the video clips from the datasets, we use the t-SNE plot of the representational embeddings of 1000 video clips from each dataset via a standard Kinetics-400~\cite{Kay2017TheKH} pretrained model. The t-SNE visualization highlights conspicuous overlap among the datasets, suggesting that their feature representations share common latent structures. While DoTA exhibits some degree of separability, CCD and DAD are more intertwined, indicating a higher degree of similarity. This overlap underscores the challenges of cross-dataset generalization and emphasizes the importance of robust domain adaptation techniques in Figure \ref{fig:tsne}.

\subsection*{A.3 Additional Results}

\begin{table}[t]
    \centering
    \setlength{\tabcolsep}{5pt}{%
    \begin{tabular}{l c c c c c c c}
        \hline
        \textbf{ID} & \textbf{MS} & \textbf{T} & \textbf{S} & \textbf{NS} & \textbf{CCD} & \textbf{DoTA} & \textbf{DAD} \\ \hline 
        1 & \cmark & \cmark & \cmark & \cmark & \textbf{0.366} & \textbf{1.747} & \textbf{0.705} \\ \hline
        2  & - & \cmark & \cmark & \cmark & 0.494 & 1.804 & 0.737 \\         
        3 & \cmark & - & \cmark & \cmark & 0.391 & 1.759 & 0.722 \\ 
        4 & \cmark & \cmark & - & \cmark & 0.394 & 1.748 & 0.727 \\ 
        5 & \cmark & \cmark & \cmark & - & 0.415 & 1.765 & 0.728 \\ 
        6  & - & \cmark & \cmark & - & 0.945 & 1.774 & 0.733 \\ 
        7  & \cmark & - & \cmark & - & 0.458 & 1.777 & 0.723 \\ 
        8  & \cmark & - & - & \cmark & 0.442 & 1.768 & 0.721 \\ 
        9 & \cmark & \cmark & - & - & 0.386 & 1.756 & 0.731 \\ 
        10  & - & - & - & \cmark & 0.534 & 1.894 & 0.773 \\ 
        11  & - & - & \cmark & - & 1.092 & 1.822 & 0.746 \\ 
        12  & - & \cmark & - & - & 0.459 & 1.783 & 0.728 \\ 
        13  & \cmark & - & - & - & 0.450 & 1.776 & 0.726 \\ 
        14  & - & - & - & - & 1.094 & 2.112 & 0.862 \\ 
        \hline
    \end{tabular}
    }
    \caption{MSE results for ablation experiments across CCD, DoTA and DAD datasets (lower is better). \textit{Multi-scale (MS)} indicates if a multi-scale structure is used in the spatial and temporal dimensions, \textit{Trend (T)} and \textit{Seasonality (S)} represent whether the disentanglement of trend and seasonality components, were incorporated in the temporal dimension, and \textit{Non-stationarity (NS)} indicates if disentanglement of non-stationary information was modeled and incorporated.}
    \label{tab:methods_ablation}
\end{table}

\subsection*{ A.3.1 Ablation study}
We present the results of our ablation study in Table \ref{tab:methods_ablation}, in which we systematically remove key elements of our method. We begin by removing the multi-scale architecture (ID 2), which results in a drastic performance drop compared to the full model (ID 1). Next, removal of other individual components (ID 2, 3, 4) results in various amounts of degradation. 
Furthermore, we observe that the removal of non-stationarity modeling (ID 5) causes more pronounced degradation than the removal of trend (ID 3) and seasonality (ID 4) pattern modeling. 
While removing two components highlights their complementary roles in performance, we notice a similar pattern in which models excluding multi-scale architecture (ID 6) and non-stationarity modeling exhibit more significant drops than all other tests in which two components are removed (ID 7, 8, 9). The smallest performance degradation occurs when seasonality and non-stationarity modeling are both removed (ID 9).
 
Next, we remove three components at a time, where we observe that ablating all components except for the seasonality component (ID 11) results in the largest degradation compared to the removal of other components (ID 10, 12, 13), suggesting that seasonality has the least individual contribution to the overall performance. 
Additionally, the degradation is the least when all the components except for the multi-scale structure are removed, further highlighting its effectiveness. 
Finally, removing all components (ID 14) expectedly leads to the worst performance across the datasets. We also observe the synergy between the multi-scale structure and other components, as models containing multi-scale architectures with at least one additional component in most cases experience less degradation (e.g. ID 6 vs. ID 7-9). This is expected, as the multi-scale structure allows for disentanglement to occur across each scale, facilitating robust video representation learning and contributing to an overall improvement in TTC performance.

\subsection*{A.3.2 Sensitivity analysis}
\label{subsec:hypersensitivity_training}

\begin{figure}[t]
  \centering
    \centering
    \includegraphics[width=1\columnwidth]{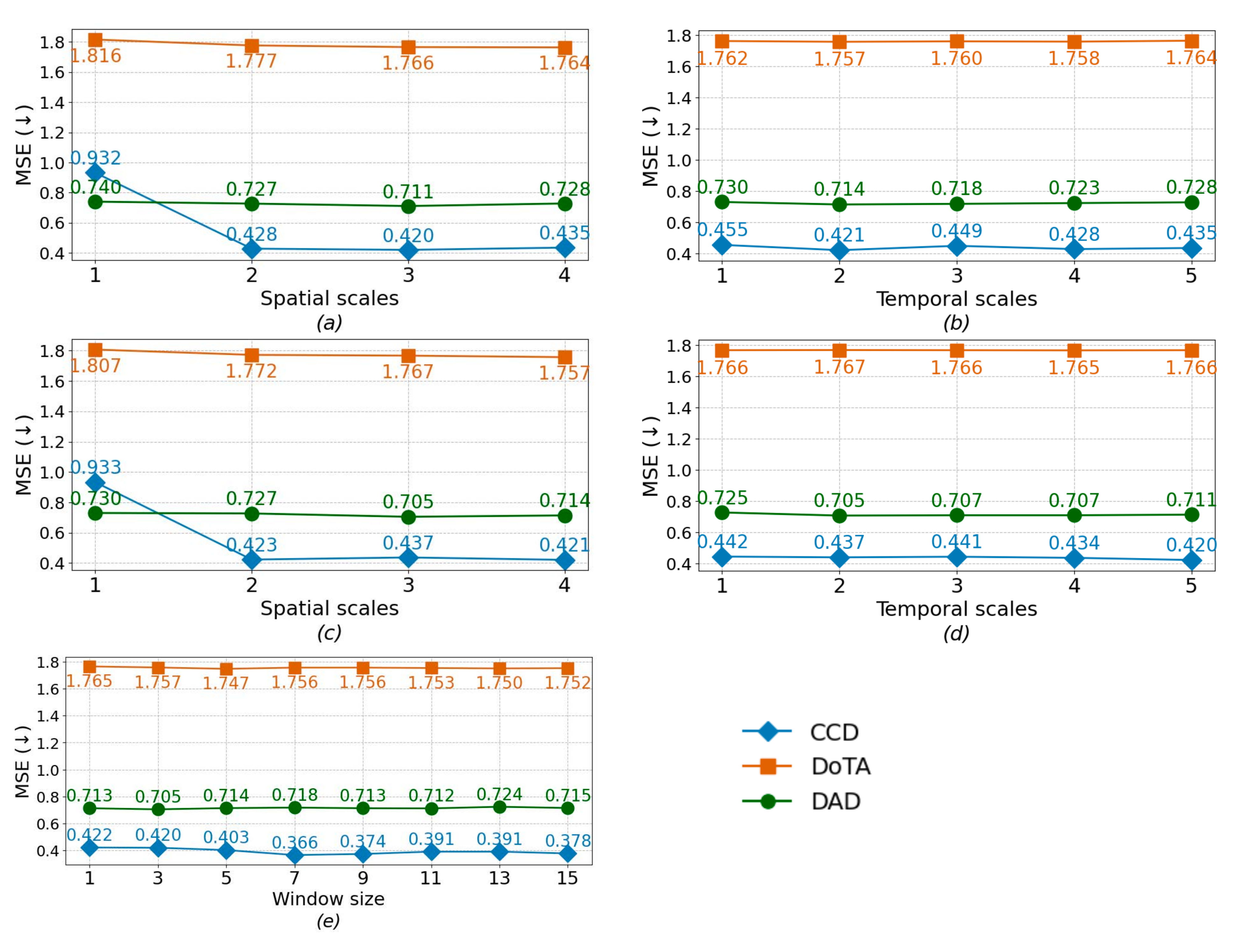}
    \caption{Sensitivity analysis for the effect of spatial scales, temporal scales, and decomposition window size.
    }
  \label{fig:hyper_sensitivity}
\end{figure}

\noindent\textbf{Impact of spatial multi-scale structure.}
We assess the sensitivity of the results to the number of spatial scales in CollideNet while keeping the temporal scales fixed at five, the maximum value, across all three datasets, as shown in Figure \ref{fig:hyper_sensitivity}(a). Across all datasets, utilizing more than one spatial scale improves performance, indicating the benefits of multi-scale spatial encoding. However, no further gains are observed beyond three scales, and using four scales results in performance degradation. Additionally, we find that the CCD dataset is the most sensitive to variations in spatial scales compared to the other datasets.

\begin{table*}[t]
    \centering
\setlength{\tabcolsep}{5pt}    
\resizebox{\linewidth}{!}{%
    \begin{tabular}{l c c c c c c}
        \hline
        \textbf{Dataset} 
        & \textbf{ADF Stat} 
        & \textbf{ADF \textit{p}-value} 
        & \textbf{KPSS \textit{p}-value}
        & \textbf{ADF Stat (norm)} 
        & \textbf{ADF \textit{p}-value (norm)} 
        & \textbf{KPSS \textit{p}-value (norm)}  
        \\ \hline 
        DAD & -2.848 & 0.170 & 0.048 & -4.734 & 0.005 & 0.100 \\
        CCD & -2.323 & 0.240 & 0.048 & -3.899 & 0.011 & 0.100 \ \\
        DoTA & -2.531 & 0.199 & 0.037 & -4.898 & 0.002 & 0.100 \ \\
         \hline
    \end{tabular}
    }
    \caption{Analysis of non-stationarity in the video frame embeddings.}
    \label{tab:dataset_stationarity}
\end{table*}

\noindent\textbf{Impact of temporal multi-scale structure.}
In Figure \ref{fig:hyper_sensitivity}(b), we evaluate the effect of varying temporal scales on performance while setting the spatial scale at four. We find a similar pattern to that of spatial scales, where using more than one scale or adopting the multi-scale architecture improves performance. Moreover, using two scales gives the best results, after which performance slightly declines across all datasets. 
In comparison, our method shows more resilience to the number of temporal scales than the spatial scales.

\noindent\textbf{Further analysis of mixed scales.}
We further examine the sensitivity by fixing one dimension at its optimal value and varying the other. Figure \ref{fig:hyper_sensitivity}(c) shows the performance of different spatial scales with the temporal scale fixed at two scales. We observe a near-identical pattern as that when fixing the temporal scale at its maximum value in Figure \ref{fig:hyper_sensitivity}(a); however, we notice that the performance of CCD and DoTA is best at the maximum temporal scale, i.e., four scales. 
In Figure \ref{fig:hyper_sensitivity}(d), the sensitivity to different temporal scales with the spatial scale fixed at three is shown. We observe higher dataset-specific sensitivity in temporal scales, where CCD and DoTA perform better at higher temporal scales, while DAD performs best at two temporal scales.

\noindent\textbf{Impact of decomposition window size.}
Figure \ref{fig:hyper_sensitivity} (e) examines the effect of the decomposition window size ($k$) in the temporal modeling of our method.
While the plot indicates high resilience of model performance to changes in window size, we observe that temporal smoothing improves performance compared to no smoothing ($k=1$). The optimal window size varies: $k=7$ for CCD, $k=15$ for DoTA, and $k=3$ for DAD, highlighting dataset-specific sensitivities to smoothing.

\subsection*{A.3.3 Non-stationarity}
We conduct an extensive study to validate the presence of non-stationarity in the datasets and further assess the impact 
of attenuating the non-stationarity caused by statistical moments (mean and variance). 
First, we leverage pretrained frozen CLIP-ViT~\cite{Radford2021LearningTV} to extract frame-level embeddings. Next, we treat these embeddings as multivariate time-series and use the Augmented Dickey-Fuller (ADF)~\cite{ADFTest} and the KPSS~\cite{KWIATKOWSKI1992159} tests to determine their non-stationarity. 
These tests provide both a test statistic and a \textit{p}-value, allowing us to assess the presence of non-stationarity from complementary perspectives.
As observed in Table \ref{tab:dataset_stationarity}, all three datasets exhibit an ADF \textit{p}-value $\geq 0.05$ and a KPSS \textit{p}-value $< 0.05$, indicating strong evidence of non-stationarity in the extracted time-series. 
Next, we intend to evaluate if the normalization module in the Non-stationarity Modeling component of our method has been successful in stationarizing the embedded video frames.
Table \ref{tab:dataset_stationarity} shows the results of the stationarity tests on the previously used CLIP embeddings, this time after applying the same normalization step used in our model.
We observe lower ADF and higher KPSS \textit{p}-values, which point to an increased amount of stationarity, indicating that our method successfully reduces non-stationarity in the embeddings, thus allowing for better TTC forecasting with transformers.

\subsection*{A.3.4 Visualization of trend and seasonality}
\label{subsec:viz_for_trend_and_seasonality}

\begin{figure}[h]
  \centering
  \begin{subfigure}[t]{\linewidth}
      \centering
      \includegraphics[page=1,width=0.9\linewidth]{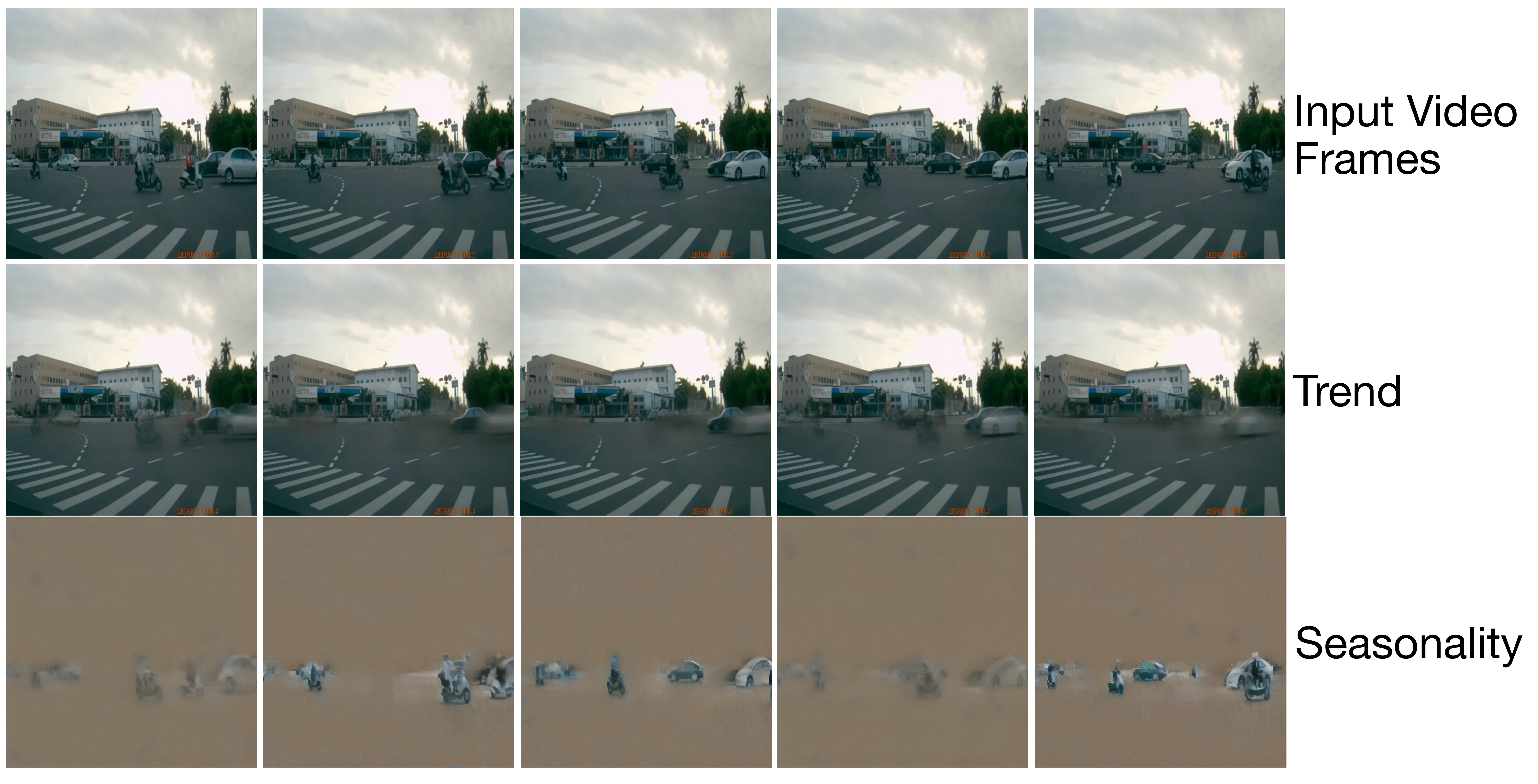}
      \caption{Static background}
  \end{subfigure}
  \begin{subfigure}[t]{\linewidth}
      \centering
      \includegraphics[page=2,width=0.9\linewidth]{assets/reconstructed_frames.pdf}
      \caption{Dynamic background}
  \end{subfigure}
  
  \caption{Auto-encoder reconstruction of trend and seasonality components in (\textit{a}) static background and  (\textit{b}) dynamic background.}
  \label{fig:recon}
\end{figure}

To gain deeper insights into the underlying trend and seasonality components of the video frame embeddings, we employ a pre-trained image autoencoder~\cite{von-platen-etal-2022-diffusers} to reconstruct these components, as shown in Figure \ref{fig:recon}. The figure illustrates two samples, (a) with a static background, and (b) with a dynamic background. We observe that in both samples, the trend component predominantly captures the background and the overall motion flow of objects, while the seasonality component emphasizes short-range and finer details, such as vehicles and traffic signs. This example highlights our method's ability to effectively disentangle trend and seasonality components, as further supported by the ablation study, where their removal significantly impacts TTC forecasting performance. 

\end{document}